\documentclass{article}

\usepackage[dblblindworkshop, final]{neurips_2026}

\usepackage[utf8]{inputenc}
\usepackage[T1]{fontenc}
\PassOptionsToPackage{hyphens}{url}
\usepackage{hyperref}
\usepackage{booktabs}
\usepackage{amsmath}
\usepackage{microtype}
\usepackage{placeins}
\usepackage{xcolor}      
\usepackage{graphicx}
\graphicspath{{./}{figures/}{images/}{figs/}}
\usepackage{tabularx}

\newcommand{\auc}{\mathrm{AUC}}
\newcommand{\arm}{\texttt{Arm}}
\newcommand{\ant}{\texttt{Omnidirectional Ant}}
\newcommand{\walker}{\texttt{Walker 2D}}

\title{From Admission to Retention: Predicting Durable Offspring in Quality-Diversity Search}

\workshoptitle{Self-Evolving Diversity-Driven Search
for Robust AI Systems (EvoRobust)}

\author{%
\begin{tabular}{@{}c@{\qquad}c@{}}
\textbf{Corrado Calamita}%
\thanks{These authors contributed equally and share first authorship.}
&
\textbf{Matteo Mendicini}\footnotemark[1]
\\
\textbf{Luca Ferrari}%
\thanks{These authors contributed equally in supporting roles.}
&
\textbf{Riccardo Bertamini}\footnotemark[2]
\end{tabular}
\\[1em]
BAINSA -- Board of AI and Neuroscience Student Association
}

\begin{document}

\maketitle

\begin{abstract}
Automated failure discovery can be framed as a diversity problem: the objective is to expose
distinct ways a system fails rather than a single critical adversarial case. Yet uniform parent
sampling provides no way to allocate search effort toward individuals likely to produce durable
archive contributions. We study this credit assignment problem in an unstructured
Quality-Diversity (QD) archive, testing whether signals available before an individual has
reproduced distinguish offspring that merely enter the archive from those still retained
generations later. Across a planar-arm benchmark and two robotic-locomotion tasks, the same
signals are often associated with entry and persistence in opposite directions. In particular,
lineage depth is the clearest such signal, and its persistence discrimination grows with
population capacity, which is associated with lower turnover. Biasing parent selection toward the signals
with the strongest measured association to survival nevertheless does not improve search:
moderate bias matches uniform sampling, and strong bias reduces QD-score.
\end{abstract}

\section{Introduction}

Failure discovery can require a collection of behaviourally distinct failures. This objective has motivated QD-based scenario generation \citep{fontaine2021hri}. QD algorithms aim to produce collections of behaviorally diverse and high-performing solutions \citep{lehman2011localcompetition,pugh2016quality}, specifically MAP-Elites \citep{mouret2015illuminating} uses a fixed descriptor grid and CVT-MAP-Elites still partitions a bounded descriptor space in advance \citep{vassiliades2018cvt}. DNS \citep{bahlousboldi2025dns} differs from this perspective: it removes the partition while retaining a descriptor, distance metric, and finite population capacity. We use that setting as motivation and evaluate on standard QD benchmarks, where behaviour spaces are predefined enabling the focus to be the measurement of parent offspring events. 
In typical MAP-Elites implementations, pressure toward exploitation acts mainly during replacement in cells while retained elites are sampled uniformly as parents \citep{mouret2015illuminating}. Related selection pressures appear in different QD frameworks. Cully and Demiris use accumulated offspring success to bias archive-parent selection \citep{cully2018quality}, and Wang et~al.\ rank MAP-Elites parents using non-surrounded dominance
\citep{wang2023nsds}, observing that QD algorithms ``often select the parent solutions
uniformly at random, which lacks selection pressure and may limit the performance''.

The same question, being how to allocate search effort from accumulated performance
information rather than uniformly, has a longer history in evolutionary
computation. Most directly, Sfikas et al. formulate MAP-Elites parent selection as a multi-armed bandit problem and use the survival rate of a parent’s offspring as an indirect reward \citep{sfikas2021mce}. That signal provides accumulated feedback for previously selected parents; our question instead asks whether useful predictors exist before a parent has produced any offspring.

Such feedback is informative only after an individual has reproduced.
\citet{kistemaker2011critical} formalise the potential of a genotype as its
descendant fitness probability (DFP), the distribution over the fitness of its
future descendants, and use it to analyse behaviour characterisations in novelty search
rather than as a selection rule. They are explicit that it cannot serve as one, for two
reasons: even a simplified DFP is intractable on non-trivial problems, and were it known,
the descendants' fitnesses would be known with it, leaving no optimisation problem to
solve. Any practical rule must therefore substitute a proxy that is cheap and already
available, and the empirical question is which proxies carry information.

\section{Setting and Method}

\paragraph{Dominated Novelty Search.}
Let $\mathcal{P}=\mathcal{A}\cup\mathcal{O}$ be the pool formed by the current archive
$\mathcal{A}$ and the offspring batch $\mathcal{O}$, and let $f_i$ and $d_i$ denote the
objective fitness and the descriptor of individual $i$. DNS assigns each individual a
competition fitness: its mean descriptor-space distance to the $k=3$ nearest
individuals in $\mathcal{P}$ of equal or higher objective fitness. Writing
$N_i=k\text{-}\mathrm{NN}\!\left(\left\{\,j\in\mathcal{P},\ j\neq i\ :\ f_j\geq f_i\,\right\}\right)$
for that neighbour set,
\begin{equation}
\widetilde f_i=\frac{1}{|N_i|}\sum_{j\in N_i}\lVert d_i-d_j\rVert_2 .
\label{eq:dns}
\end{equation}
If fewer than $k$ such individuals exist we average over those that do; if there are none
we set $\widetilde f_i=+\infty$.
The pooled set is then ranked by $\widetilde f$ from highest to lowest and truncated to the population capacity. Distances use unnormalised Euclidean distance on the raw descriptors. Two consequences follow. Eviction is global and recurrent rather than
local and triggered: each archive update re-ranks the entire pooled set, so an
individual can be displaced at any later generation by an event anywhere in
descriptor space. Retention is therefore an ongoing outcome rather than a
one-time contest for a cell, which is what makes persistence worth measuring
separately from entry, and makes archive age and lineage history plausible signals of repeated global competition. Despite having several archive and lineage signals available, DNS still samples parents uniformly. This leaves open whether any signal available before reproduction can identify parents whose offspring will persist.

\paragraph{Signal availability.}
For a signal to guide an individual’s first selection as a parent, it must be defined before that individual has produced offsprings. 
Cold-start selectability is distinct from being fixed at birth: age is selectable before reproduction although it changes over time. Lineage depth is the number of reproduction steps from the founder along the base-parent chain and is fixed at creation. Prior QD work has related genealogical diversity to performance post hoc \citep{nordmoen2021stepping}; here we test lineage information as a candidate selection-time signal. The archive-state group provides our contemporaneous baselines: the selection quantities
the literature already proposes, recomputed from the current archive at every generation.

\begin{table}[t]
\caption{Signal availability.}
\label{tab:taxonomy}
\centering
\scriptsize
\setlength{\tabcolsep}{3pt}
\begin{tabularx}{\linewidth}{@{}p{0.20\linewidth}cX@{}}
\toprule
Group & Cold-start? & Signals \\
\midrule
Lineage at birth & yes & depth; ancestral-fitness EWMA; ancestry entropy; competition fitness at birth \\
Current state & yes & age \\
Archive state & yes & parent fitness; current competition fitness \\
Mating event & no & $|\lambda|$; compositional residual; child fitness \\
Offspring feedback & no & curiosity; offspring survival rate \\
\bottomrule
\end{tabularx}
\end{table}

Feedback signals remain useful diagnostics: at capacity 1024, on \arm{}, \ant{} and
\walker{} respectively, curiosity reaches entry AUC 0.738, 0.603 and 0.601, and offspring
survival rate 0.673, 0.581 and 0.564. They are nevertheless undefined for a newcomer.

\paragraph{Protocol.}
We distinguish between two moments: immediate entry into the archive and long-term persistence. We measure all twelve signals in Table~\ref{tab:taxonomy} (eleven at capacity 4096; see Appendix~A.2). All AUC measurements are made under uniform parent selection.
Entry means that an offspring is kept in the archive immediately after it is created. Persistence means that it is still in the archive $H$ generations later. We use $H=1000$ for the main results and check $H\in\{100,1000\}$ on all tasks and $H=10$ on \arm{}, where archive logging is per-generation; signs are stable across horizons. Offspring created too close to the end of a run to measure persistence are excluded.

Let $g^{+}$ denote the signal value of a parent whose offspring achieved the outcome and
$g^{-}$ that of a parent whose offspring did not. For each signal we compute
\begin{equation}
\auc=\Pr\!\left(g^{+}>g^{-}\right)+\tfrac{1}{2}\Pr\!\left(g^{+}=g^{-}\right),
\qquad
\Delta=\left|\auc-0.5\right| ,
\label{eq:auc}
\end{equation}
and report the discrimination strength $\Delta$ together with the sign of $\auc-0.5$. An
$\auc$ of 0.5 means no discrimination; $\auc<0.5$ means the association runs in the opposite
direction, which does not imply that inverting the signal yields a good selection rule.
We compute one AUC per signal per run, with different seeds per (domain, capacity) cell (Table~\ref{tab:runs}), and use runs, rather than individual parent-offspring pairs, as the unit of analysis. This avoids treating the ${\approx}2.5\times10^{7}$ correlated pairs as independent samples.

We evaluate three domains at archive capacities 256, 1024 and 4096, holding offspring per
generation fixed at 256, so the ratio of new candidates to available slots is 1.00, 0.25 and
0.06 and larger archives turn over more slowly. Each measurement run lasts 10,000
generations with $k=3$ and uniform parent sampling. Measured across domains, discrimination
and archive entry rate co-vary: entry rates are approximately 0.13 for \arm{}, 0.10 for
\ant{} and 0.04 for \walker{}. Varying capacity within a fixed domain holds the operator, the
descriptor space and the per-generation evaluation budget identical, but it is not a clean
instrument for turnover alone: it also changes the offspring-to-slot ratio, the competition
geometry and the attainable coverage. We therefore report an association between capacity and
discrimination that is consistent with a turnover mechanism, rather than an isolated causal
effect of turnover.

\paragraph{Selection rule.}
For the intervention we replace uniform sampling of the base parent with a rank-softmax,
$P(i)\propto\exp(\mathrm{rank}(g_i)/\tau)$, over the occupied slots, so the ratio between the
most- and least-favoured parent is $\exp(1/\tau)$ whatever the signal's units; ranks are
tie-averaged for integer-valued signals such as depth. Sweeps are listed in
Table~\ref{tab:runs}.

\section{Results}

\begin{figure}[htb!]
\centering
\includegraphics[width=0.9\linewidth,height=0.24\textheight,keepaspectratio]{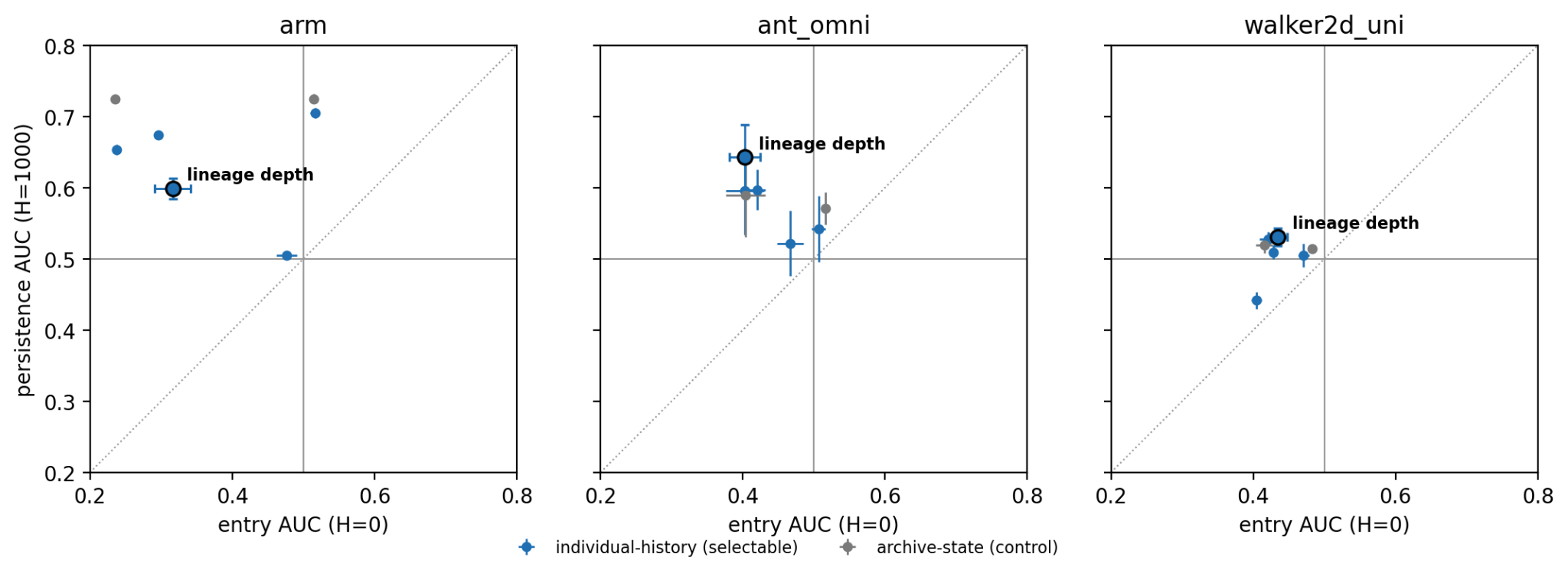}
\caption{Entry against persistence AUC at capacity 1024 and $H=1000$, one panel per domain.
Points are cold-start-selectable signals averaged across seeds; blue denotes lineage- and
current-state signals, grey the contemporaneous archive-state baselines. Grey lines mark
chance. Signals on opposite sides of chance for the two outcomes are associated with entry
and persistence in opposite directions.}
\label{fig:entry-persistence}
\end{figure}

\paragraph{Entry is not persistence.}
At capacity 1024, several selectable signals reverse direction between immediate entry and persistence after $H=1000$ generations on all three tasks (Figure~\ref{fig:entry-persistence}). This means that properties associated with easier immediate admission can be associated in the opposite direction with survival 1000 generations later. Lineage depth is the clearest case: shallower lineages are associated with entry, while deeper lineages are associated with persistence.

This describes a rank association and does not show which signal, if any, would improve coverage when used for selection.

\begin{figure}[htb!]
\centering
\includegraphics[width=0.9\linewidth,height=0.20\textheight,keepaspectratio]{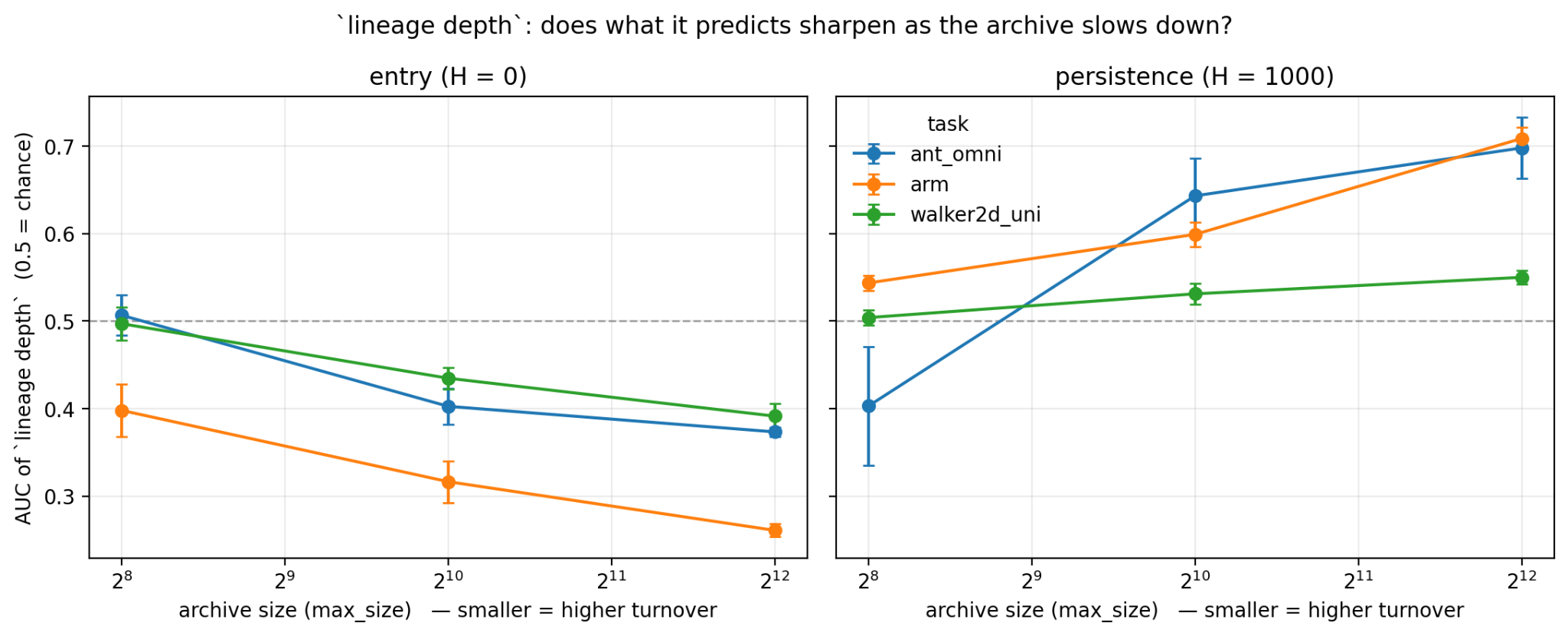}
\caption{Signed lineage-depth AUC across archive capacities for entry and persistence. Larger archives have lower turnover because the number of offspring stays fixed.}
\label{fig:turnover}
\end{figure}

\paragraph{Turnover and depth spread.}
The weaker AUC at high turnover is not caused by retained individuals having very similar lineage depths. In fact, depth variation is already large at capacity 256 and usually decreases as capacity grows. For example, on \ant{}, persistence discrimination increases from 0.097 to 0.198 (Figure~\ref{fig:turnover}) while the interquartile range of lineage depth drops from ${\approx}94$ to ${\approx}16$. The association strengthens on \arm{} and \walker{} as well, without any comparable narrowing of depth. Lineage-depth persistence AUC rises monotonically with population capacity on all three tasks, while entry AUC falls below chance (Figure~\ref{fig:turnover}).

\paragraph{Neither signal converts into a selection rule.}
The turnover result makes a falsifiable prediction. Parent fitness discriminates
persistence on \ant{} roughly five times as well at capacity 4096 as at 256
(from 0.027 to 0.138), so if biasing toward it fails at low
capacity for want of association to exploit, it should succeed at high capacity.
We found it does not. Moderate pressure ($\tau=0.5$) matches uniform DNS at both
capacities, reaching projected QD-score $4.04\times10^{6}$ against
$4.02\times10^{6}$ for uniform at capacity 4096 over 10 seeds; strong pressure
($\tau=0.1$) is substantially worse, and worse at the higher capacity
($-64.4\%$ against $-43.6\%$ at capacity 1024). This represents a loss of diversity rather than
of quality as: projected coverage drops from 0.9999 to 0.517, mean novelty drops from 0.848 to 0.544 while maximum fitness stays essentially unchanged. The birth-available signal behaves no better. On \ant{} at capacity 1024 over 10,000
generations, rank-softmax on lineage depth ($\tau=0.5$) reaches projected QD-score
$3.130\times10^{6}$ at coverage 0.767; weakening the bias ($\tau=2.0$) gives
$3.112\times10^{6}$; inverting it gives $3.125\times10^{6}$ at coverage 0.769. The matched
uniform run at the same capacity and horizon reaches $3.156\times10^{6}$ at coverage 0.773.
No depth policy exceeds uniform sampling on either metric, and every difference falls within
the between-seed spread, so favouring deep lineages, weakening that preference, and
reversing it are all indistinguishable from sampling uniformly. Coverage here is 0.77 rather
than saturated, so this null is not an artefact of a coverage ceiling.

\FloatBarrier
\section{Conclusion}

Entry and long-term persistence are associated with different, often oppositely signed parent
properties, so the two outcomes should be evaluated separately, and a signal is usable only if
it is defined before an individual has reproduced. Lineage depth satisfies both conditions and
its retention association strengthens as turnover falls, even as the spread of depth narrows,
but it does not translate into a better selection rule. These results concern predictive
association rather than causal credit, and are limited to DNS, one variation operator and
three standard QD domains.

This null is not specific to parent selection: perturbing DNS's \emph{survival} rule
degrades QD-score monotonically in the perturbation size on all three domains, again
through coverage rather than fitness, and how far a rule moves the selection ordering
predicts the loss (Appendix~\ref{app:rulespace}). DNS's ordering is hard to improve on
locally, which bounds what a different rule \emph{form} can buy. More broadly,
if turnover governs how informative lineage history is, then archive capacity is not only a
budget parameter but a choice about what credit assignment is possible at all, a
parameter a self-evolving search could adapt during a run rather than fix in advance.

\clearpage
\appendix

\section{Supplementary Details}

\subsection{Lineage-depth spread}

One possible explanation for the capacity effect is that high turnover gives archive members little time to develop different histories. If their histories remain too similar, a history-based signal could appear close to chance even if the underlying idea is useful. That story requires spread and
association to rise together, and they do not (Table~\ref{tab:depth-iqr}). Depth IQR
never falls below ${\approx}13$ over a 10,000-generation run, so the near-chance AUC at
capacity 256 is not explained by absent variation. Values are read from per-generation
archive traces and reported to two significant figures.

\begin{table}[htb!]
\caption{Interquartile range of lineage depth among retained individuals.}
\label{tab:depth-iqr}
\centering
\small
\begin{tabular}{lccc}
\toprule
Task & 256 & 1024 & 4096 \\
\midrule
\ant{}    & ${\approx}94$ & ${\approx}32$ & ${\approx}16$ \\
\arm{}    & ${\approx}16$ & ${\approx}15$ & ${\approx}13$ \\
\walker{} & ${\approx}22$ & ${\approx}23$ & ${\approx}16$ \\
\bottomrule
\end{tabular}
\end{table}

\subsection{Runs}

All conditions use $k=3$, 256 offspring per generation, and Iso+LineDD variation
\citep{vassiliades2018elite} with $\sigma_{\mathrm{iso}}=0.005$ and
$\sigma_{\mathrm{line}}=0.05$.

\begin{table}[htb!]
\caption{Experimental configuration. Seed counts differ by capacity in the
measurement sweep.}
\label{tab:runs}
\centering
\scriptsize
\setlength{\tabcolsep}{4pt}
\begin{tabular}{llllll}
\toprule
Experiment & Domains & Capacity & Gens & Parent selection & Seeds \\
\midrule
Signal measurement & all three & 256 & 10,000 & uniform & 5 \\
                   & all three & 1024 & 10,000 & uniform & 10 \\
                   & all three & 4096 & 10,000 & uniform & 5 \\
Fitness sweep & \ant{} & 1024 & 4,000 & uniform, $\tau\in\{0.1,0.5\}$ & 10 \\
              & \ant{} & 4096 & 4,000 & uniform, $\tau\in\{0.1,0.5\}$ & 10 \\
Depth sweep & \ant{} & 1024 & 10,000 & $\pm$depth, $\tau\in\{0.5,2.0\}$ & 10 \\
\bottomrule
\end{tabular}
\end{table}

Two instrumentation limits apply. The Brax domains log the archive every tenth
generation, so the $H=10$ persistence horizon is undefined there and is reported
only for \arm{}. Ancestry entropy is undefined at capacity 4096 on all three
tasks for memory reasons, so that condition carries 11 of the 12 signals.

Domain specifications: \arm{} uses 128 joint angles in $[0,1]$ with an
end-effector descriptor in $[0,1]^2$ and objective $-\mathrm{std}(\text{angles})$;
\ant{} uses Brax spring physics, final position in
$[-30,30]^2$ as descriptor, forward reward removed, horizon 1000; \walker{} uses
Brax spring physics, an MLP [16] policy, a feet-contact descriptor in $[0,1]^2$,
the standard Walker2d reward, and horizon 1000.

\subsection{Where the null comes from: departures from DNS's ordering are costly}
\label{app:rulespace}

A companion sweep perturbs the survival rule rather than the parent-selection rule, on the
same three domains and the same archive (capacity 1024, 256 offspring per generation,
$k=3$, uniform parent selection; 10,000 generations and 10 seeds on \arm{}, 4,000
generations and 5 seeds on the Brax domains). The perturbation mixes a second notion of crowding, distance to a bounded record of previously evaluated individuals, into the competition score with weight $\lambda$. So $\lambda=0$ recovers
DNS exactly. The ladder is monotone decreasing on every domain, so the optimum is the
endpoint rather than an interior mixture.

\begin{table}[htb!]
\caption{Final projected QD-score as a percentage of unmodified DNS ($\lambda=0$). The
ladder is monotone on every domain.}
\label{tab:rulespace}
\centering
\small
\begin{tabular}{lccc}
\toprule
Domain & $\lambda=0.25$ & $\lambda=0.50$ & $\lambda=1.00$ \\
\midrule
\arm{}    & 98.9 & 96.0 & 66.0 \\
\ant{}    & 97.0 & 94.2 & 78.5 \\
\walker{} & 95.8 & 94.6 & 60.9 \\
\bottomrule
\end{tabular}
\end{table}

The ranking of rules replicates across domains (Kendall $\tau$ between domain rankings 0.81
to 0.90, with DNS first and $\lambda=1$ last on all three). Within each domain, the Spearman
correlation between a rule's ordering of candidates and DNS's ordering predicts its final
QD-score: $r=0.89$ on \arm{}, 0.95 on \ant{} and 0.91 on \walker{}, and 0.90 over all 21
domain-rule pairs. The further a rule moves the selection ordering, the more it loses.

The failure mode matches the one reported in the main text. Over all runs at the final
generation, projected coverage correlates with projected QD-score at 0.998 on \ant{} and
0.954 on \walker{}, while maximum fitness correlates at $-0.39$ and 0.06. On \arm{} the two
are not independent evidence, since with the offset used there QD-score equals coverage to
a relative error of $10^{-7}$.

\end{document}